# Don't miss the mismatch: investigating the objective function mismatch for unsupervised representation learning

Bonifaz Stuhr[1,2] · Jürgen Brauer[2]

## Abstract

Finding general evaluation metrics for unsupervised representation learning techniques is a challenging open research question, which recently has become more and more necessary due to the increasing interest in unsupervised methods. Even though these methods promise beneficial representation characteristics, most approaches currently suffer from the objective function mismatch. This mismatch states that the performance on a desired target task can decrease when the unsupervised pretext task is learned too long–especially when both tasks are ill-posed. In this work, we build upon the widely used linear evaluation protocol and define new general evaluation metrics to quantitatively capture the objective function mismatch and the more generic metrics mismatch. We discuss the usability and stability of our protocols on a variety of pretext and target tasks and study mismatches in a wide range of experiments. Thereby we disclose dependencies of the objective function mismatch across several pretext and target tasks with respect to the pretext model's representation size, target model complexity, pretext and target task augmentations as well as pretext and target task types. In our experiments, we find that the objective function mismatch reduces performance by $\sim 0.1$–$5.0\%$ for Cifar10, Cifar100 and PCam in many setups, and up to $\sim 25$–$59\%$ in extreme cases for the 3dshapes dataset.

## 1 Introduction

Unsupervised Representation Learning is a promising approach to learn useful features from huge amounts of data without human annotation effort. Thereby, a common evaluation pattern is to train an unsupervised pretext model on different datasets and then test its performance on several target tasks. Because of the huge variety of target tasks and preferred representation characteristics, the evaluation of these methods is challenging. In recent work, a large number of evaluation metrics have been proposed [24, 39, 40, 45], but because of the fast changes in unsupervised learning methodologies only a few of them can be used across the wide spectrum of promising approaches. This is one reason why the linear evaluation protocol is now commonly used [9, 14, 16, 21, 32, 37, 46], which trains a linear model for a target task on-top of the representations of an unsupervised pretext model. In this work, we show that simply training a target model for different layers of the pretext model does not yield the entire picture of the training process and leads to a loss of useful temporal information about learning. It is already known in literature, that succeeding in a pretext task can be the reason why the model fails on the target task. Here we propose that the linear evaluation protocol does not capture this properly. Therefore, we extend this protocol and address the question of when succeeding in a pretext task hurts performance and how much. We train target models on representations obtained from different training steps or epochs of the pretext model and plot target and pretext model metrics in comparison, as shown in Fig. 1. Thereby we observe that training an unsupervised pretext model too long can lead to an objective function mismatch [41, 58]

✉ Bonifaz Stuhr
  bonifaz.stuhr@hs-kempten.de

  Jürgen Brauer
  juergen.brauer@hs-kempten.de

[1] Department of Computer Science, Autonomous University of Barcelona, 08193 Bellaterra, Barcelona, Spain

[2] Department of Computer Science, University of Applied Sciences Kempten, 87435 Kempten, Germany

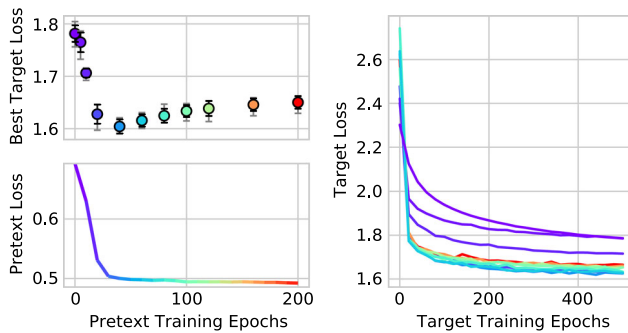

**Fig. 1** (Bottom left) Evaluation loss of a pretext autoencoder trained on Cifar10. (Top left) Best evaluation losses of linear target models trained for classification on the representations of the pretext autoencoder from different pretext training epochs. (Right) Evaluation loss curves from each linear target model. Colors correspond to the current epoch of pretext task training, and each value is obtained by 5-fold cross-validation. An objective function mismatch occurs around pretext training epoch 40, from which the target loss increases.

between the objectives used to train both models. This mismatch leads to a drop in performance on the target task, while the pretext model and the target models still converge correctly, which can be seen in Fig. 1. To quantify our results, we define soft and hard versions of two simple and general evaluation metrics - the *metrics mismatch* and the *objective function mismatch* - formally. With these metrics, we then evaluate different image-based pretext task types for self-supervised learning by using the linear evaluation protocol.

Our contributions can be summarized as follows:

- We propose hard and soft versions of general metrics to measure and compare mismatches of (unsupervised) representation learning methods across different target tasks (Sect. 3 and 4). To the best of our knowledge, this has not been done before.
- We discuss the usability and stability of our protocols on a variety of pretext and target tasks (Sect. 6.2).
- In our experiments we qualitatively show dependencies of the objective function mismatch with respect to the pretext model's representation size (Sect. 6.3), target model complexity (Sect. 6.4), pretext and target augmentations (Sect. 6.5) as well as pretext and target task types (Sect. 6.6).
- We find that the objective function mismatch can reduce performance on various benchmarks. Specifically, we observe a performance decrease by $\sim 0.1$–5.0% for Cifar10, Cifar100 and PCam, and up to $\sim 25$–59% in extreme cases for the 3dshapes dataset (Sect. 6).

# 2 Related work

## 2.1 Unsupervised representation learning

Many unsupervised representation learning algorithms are based on self-supervised learning [26, 52, 53], which obtains labels directly from data without human annotation to define a pretext task. There are several approaches to self-supervision:

*Generation-based self-supervison* examines the generation of an arbitrary output from a learned representation of the given input. One line of work improves on autoencoders [51] and variational autoencoders [31] by defining generation-based pretext tasks which lead to representations valuable for required target tasks (e.g., object classification or detection). Examples are denoising [8, 61], colorization [34, 35, 73], or inpainting of images [47, 67]. Recently, a second line of work based on GANs [17] emerged, which adjusts their latent space for representation learning, for example by constraining [50] or changing [14] the architecture. In a third line of research, an autoregressive, transformer-based model achieved state-of-the-art performance on visual representation learning by sequential image generation [20]. Generation-based self-supervison is applied to other modalities as well, e.g., audio [20] or video [57, 62].

*Context-based self-supervison* recently has moved more and more away from autoencoding data: Early approaches utilize spatial context structure by defining pretext tasks for context generation, like image inpainting [47] or denoising, as a weak form of inpainting [8, 61]. In contrast, approaches for context prediction do not create any image and, for example, try to leverage the knowledge obtained by predicting patch positions [12, 43]. Spatial context can also be encoded by predicting transformations, which has led to a line of research focusing on autoencoding transformations rather than data [16, 37, 49, 72]. Recently, the context-based similarity approach of contrastive learning [19], which utilizes context information between negative and positive pairs, gained popularity and achieved promising results [9, 11, 21, 36, 56]. Contrastive Learning has been linked to mutual information maximization [38, 68], which in ongoing work is used to define pretext tasks through context-based similarity as well [4, 24]. Context similarity by pseudo-labeling through clustering methods is another line of research [7, 71]. Self-supervised relational reasoning combines context-based similarity and context-based structure by discriminating how entities relate to themselves and to other entities and has also been linked to mutual information maximization [46]. Context-based approaches are applied to other data modalities as well, e.g., point clouds [69] or video [15, 28].

*Other unsupervised representation learning methods* for example combine multiple self-supervised approaches [10, 29, 48, 63], use meta learning [25, 27, 41, 54] or metric learning [6, 65] to learn unsupervised learning rules, or rely on self-organization [1, 58].

## 2.2 Analyzing unsupervised representation learning

*Changing the underlying model* is one common theme to compare different unsupervised learning techniques [18]. Here, a well known finding is that a larger representation size significantly and consistently increases the quality of the learned visual representations [32].

*Varying the amount of data samples* has led to interesting observations as well [42, 59]. For example in [3], it is shown that unsupervised learning is capable to learn features of early layers from a single image.

*Analyzing self-supervised learning across target domains* is another way to define and evaluate benchmarks for unsupervised approaches [8, 42, 64]. Zhai et al. [70] define good representations as those that adapt to diverse, unseen tasks with few examples.

Furthermore, there exist works where the underlying model, the amount of data samples and the target domain is analyzed collectively [18, 44].

*Other investigations of unsupervised learning* focus on the effect of the multitask pretext learning [13, 55], evaluate the disentanglement of representations [39], investigate the positive effects of unsupervised learning regarding robustness [23], or provide a theoretical analysis of contrastive learning [2, 66].

*The objective function mismatch in unsupervised learning* is not unknown. Some works directly or indirectly observed that learning a pretext task too long may hurt target task performance, but made no further investigations on this topic [32, 39, 64]. Other works sometimes showed performances of linear target models over training epochs, but did not examine or define the objective function mismatch in detail [58, 70]. Instead, unsupervised multi-task learning and meta learning are proposed as approaches to lower the objective function mismatch [13, 41]. In contrast, this work focuses on defining general protocols to measure mismatches of metrics over the course of pretext task training when a target task is trained on top of the pretext model's representations. To the best of our knowledge, this has not been done before. Furthermore, we highlight important properties of our evaluation protocols and interesting dependencies of the objective function mismatch.

## 3 Hard metrics mismatch

With the objective function mismatch, we want to measure the mismatch of two objectives while training a model on a (unsupervised) pretext task and using its representations to train another model on a target task. In general, we can measure the mismatch of two comparable metrics, if one metric is captured during training of a single pretext model and the other is captured for each target model fully trained on the representations of different steps or epochs of the pretext model. Two comparable metrics, for example, are classification accuracies for the pretext and target task, because they use the same measurement unit and scale. As illustrated in Fig. 2, the metric values of the target models form a curve over the course of learning. Between a metric value on this curve and the corresponding metric value on the pretext model curve, we can define the *metrics mismatch* (M3) for a certain step (or epoch) in training by calculating their distance.

More formally, let $M^P = (m_1^P, ..., m_n^P)$ denote an $n$-tuple of values from a metric used to measure pretext model $P$ for different steps $S = (s_1, ..., s_n)$. The length $n$ of the tuple is usually given by a convergence criterion $C$ on the metric of model $P$ during training. Furthermore, let $M^T = (m_1^T, ..., m_n^T)$ denote an $n$-tuple of values from a comparable metric used to measure target model $T$. $M^T$ is of the same length and order as $M^P$ and all values are calculated at the same training steps $S$ of $M^P$. Thereby the target model $T$ is fully retrained for every step $s_i$ in $S$ on the representations of model $P$ at this step before we measure $m_i^T$.

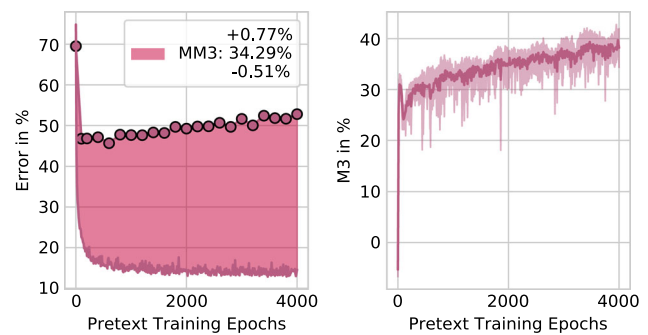

**Fig. 2** (Left) Intuition behind MM3: In this case, both metrics measure a classification error in %. The pretext metric (solid curve) is measured on the pretext task of predicting rotations with a ResNet18 model and the target metric (dotted curve) on the fully trained Cifar10 classification task. When divided by the number of measurements, the discrete area enclosed by the target and pretext task curves corresponds to the MM3 of the entire training process. (Right) Metrics Mismatch M3 plotted during training: We observe a common behavior where the mismatch increases as training progresses. Additionally, we show the stability $(+, -)$ of M3 and MM3 across a 5-fold cross-validation

**Definition 1** The hard Metrics MisMatch (M3) between $m_i^T$ and $m_i^P$ at step $s_i$ is defined as:

$$\text{M3}(m_i^T, m_i^P) := m_i^T - m_i^P \tag{1}$$

where $m_i^T$ and $m_i^P$ are single values measured with comparable metrics at step $s_i$.[1]

If M3 > 0 the performance of the target model is lower than the performance of the pretext model at step $s_i$. In contrast, M3 ≤ 0 represents the desired case in unsupervised representation learning, where target model performance is the same or above the pretext model performance at step $s_i$. In our case, we measure $m_i^T$ and $m_i^P$ over the entire evaluation dataset for every step $s_i$ in $S$. We plot $\text{M3}(m_i^T, m_i^P)$ for the pretext task of predicting rotations and the target task of Cifar10 classification during training in Fig. 2. This shows that our metric captures the behavior of the target task performance regarding the pretext task performance, and we observe an increasing mismatch as training progresses. To capture the mismatch of the entire training procedure with respect to the target task in a single value, we can now define the *mean hard metrics mismatch* (MM3) as the mean bias error between $M^T$ and $M^P$.

**Definition 2** The Mean hard Metrics MisMatch (MM3) between $M^T$ and $M^P$ is defined as:

$$\text{MM3}(M^T, M^P) := \frac{1}{n} \sum_{0 < i \leq n} (m_i^T - m_i^P) \tag{2}$$

where $M^T$ and $M^P$ are tuples measured with comparable metrics until the pretext model converges at step $s_n$.

MM3 measures the bias of the target model metric to the pretext model metric. For positive or negative values of MM3, we can make similar observations as for M3, but they now account for the tendency of the entire training process and not for a single step $s_i$. In general, the mean bias error can convey useful information, but it should be interpreted cautiously because there are special cases where positive and negative values cancel each other out. In our case, this can happen, for example, when learning the pretext task is very useful for the target task early in training but hurts the target performance equally strong later on when the pretext task is sufficiently solved. We simply capture this behavior by measuring and plotting M3 individually for the metric values of each step $s_i$ as in

Fig. 2, analogous to the way a loss is measured and plotted during training.

### 3.1 Hard objective function mismatch

Naively, we could compare the objective functions of the target and pretext task by using M3, which we define as the *hard objective function mismatch*. In most cases, however, the objective functions used to train the pretext model and the target models are not directly comparable. This is due to the usage of different objective functions for both model types, which, i.e., use different (non)linearities. But for some pretext tasks simple, comparable metrics can be defined. These metrics can be used as a proxy to measure the objective function mismatch in a general and comparable manner. A well known example is the accuracy metric, which can be used on the self-supervised tasks of predicting rotations [16] and the state-of-the-art approach of contrastive learning [9]. But comparable metric pairs can not always be found easily. For example, if we train a variational autoencoder and later use its representation for a classification target task, it does not make sense to define a pixel-wise error between the given and generated images as a comparable pretext task metric. To achieve a comparable measurement for this situation, and on the loss curves in general, one could think of individual normalization techniques between objective function pairs. However, we want to be practical and define a measure, which can be used independently of the objective function pairs for every pretext and target model combination. Furthermore, in practice, we might be especially interested in how much the target task mismatches with the pretext task if a mismatch decreases target performance. This is why we define soft versions of our measurements.

## 4 Soft metrics mismatch

To bypass objective function pair normalization, we define the *soft metrics mismatch* (SM3) directly on the target metric. Thereby, we no longer take the exact improvement of the pretext metric into account, we only care about its convergence. Since we now have no exact information about the pretext metric curve, we define SM3 for the current step $s_i$ between the current target metric value and the previously or currently occurred minimal target metric value:

**Definition 3** The Soft Metrics MisMatch (SM3) between $M^T$ and $M^P$ at step $s_i$ is defined as:

$$\text{SM3}(m_i^T) := m_i^T - \min_{0 < j \leq i}(m_j^T) \tag{3}$$

---

[1] Note that we define our measurements only for the case where lower metric values correspond to better performance. The definition for the opposite case arises naturally by changing maximum and minimum operations and/or subtraction orders.

where $\min_{0<j\leq i}(m_j^T)$ is the previously or currently occurred minimal target metric value.

SM3 has a slightly different meaning compared to M3: It equals zero if $m_i^T$ is a minimal metric value and is positive if $m_i^T$ is higher than the previously occurred minimal metric value. We want to point out that the only way we incorporate the pretext metrics into this measurement is by making sure that the pretext model does not overfit and has not yet converged. Again, we measure $m_i^T$ and $m_i^P$ over the entire evaluation dataset for every step $s_i$ in $S$ and plot SM3$(m_i^T)$. A common case is shown in Fig. 3, which captures the behavior of pretext model training with respect to the target model. Here we observe zero soft mismatch early in training followed by increasing soft mismatch until pretext model convergence. Again, we can capture the mismatch of the pretext task with respect to the target task for the entire training process until pretext model convergence as the mean bias error of every metric value $m_i^T$ and its minimal metric value:

**Definition 4** The Mean Soft Metrics Mismatch (MSM3) between $M^T$ and $M^P$ is defined as:

$$\text{MSM3}(M^T) := \frac{1}{n} \sum_{0<i\leq n} \left( m_i^T - \min_{0<j\leq i}(m_j^T) \right) \qquad (4)$$

when the pretext model convergences at step $s_n$.

MSM3 can either be zero, if no mismatch occurs, or positive, if there is a mismatch. Therefore, using MSM3 brings the benefit that positive values can not be canceled out by negative values. Furthermore, we define the maximum occurring mismatch mSM3 and the mismatch at the pretext model convergence cSM3. We are especially interested in cSM3, since it measures the representations one would naively take for the target task:

$$\text{cSM3}(M^T) := \text{SM3}(m_n^T) \qquad (5)$$

$$\text{mSM3}(M^T) := \max_{0<i\leq n} \left( \text{SM3}(m_i^T) \right) \qquad (6)$$

## 4.1 Soft objective function mismatch

Now we can use SM3 to measure a soft form of the objective function mismatch on the loss curve obtained by the target models. However, the values of these measurements lie in a range, which depends on the target objective function. Therefore, they are not directly comparable to the measurements on loss curves from other target tasks. This is why we normalize the measurements of the target metric to percentage range and define the *objective functions mismatch* (OFM) as follows:

**Definition 5** The Soft Objective Function Mismatch (OFM) between $M^T$ and $M^P$ at step $s_i$ is defined as:

$$\text{OFM}(m_i^T) := \text{SM3}\left( N(m_i^T) \right) \qquad (7)$$

$$N(x) := \begin{cases} \dfrac{100 \times x}{m_1^T - m_b^T} & m_1^T & gt; x \geq m_b^T \\ 0 & m_1^T = m_b^T = x \\ \infty\, (special case) & m_1^T = m_b^T & lt; x \end{cases} \qquad (8)$$

where $m_1^T$ is the loss value of the target model trained on an untrained pretext model ($s_1 = 0$) and $b = \operatorname{argmin}_{0<i\leq n}(m_i^T)$ denotes the index of the minimal target loss value. We then use $\hat{M}^T = (N(m_1^T), ..., N(m_n^T))$ to calculate the OFM.

The intuition behind this normalization is that we declare $m_b^T$ as the value where the pretext model has learned all of the target objective, it was able to learn (with this setting) and $m_1^T$ as the value where the model has learned nothing of the target objective. Now we measure with OFM$(m_i^T)$ for what percentage the learning of a pretext objective hurts the maximum achieved target performance at step $s_i$. An example is illustrated in Fig. 3. Furthermore, we can normalize the other soft measurements from Eqn. 5 and 6 analogs to Eq. 7.

The OFM is a general measure, which can be used for pretext and target models where no good proxy metrics can be defined. With the OFM we are able to compare mismatches across different pretext and target task objectives and their combinations. We propose these measurements to obtain quantitative and therefore comparable results for individual pretext tasks. To get the best information about the training process, we encourage to plot the curves formed by our metrics as well. We want to point out that our metrics are not intended to measure target task performance, they measure how much the performance on a target task can decrease when an (ill-posed) pretext task is learned too long. Now, to understand the OFM further, we take a look at some cases:

OFM$(m_i^T) = 0$: In this case, solving the pretext task objective did not hurt the performance of the target task objective at this point in training.

OFM$(m_i^T) = x$: Solving the pretext objective did hurt the performance of the target objective at this point in training by $x\%$ of what the model has learned. Therefore, we should have stopped training earlier. It is not guaranteed that longer training would hurt performance even more, but a growing OFM curve or MOFM is a good indicator for that.

OFM$(m_i^T) > 100$ The target objective performance is worse than for the untrained model at this point in training.

MOFM$(M^T) = \infty$ Solving the pretext objective hurts the performance of the target objective from the point of initialization. Because we have learned essentially 0% about the target objective in the training process, there is no interval to be used for normalization. Therefore, we interpret this case as if the model has an infinite mismatch as soon as the model forgets something about the target objective.

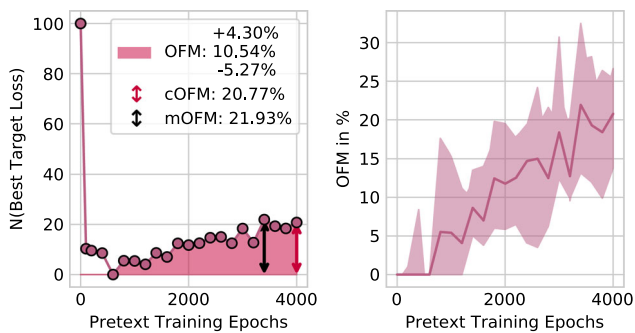

**Fig. 3** We measure the OFM instead of SM3 by normalizing the metric values with Equation 8. For visualization, we additionally shift the normalized metric values such that they lie in [0, 100] by subtracting the minimal measurement. (Left) Intuition for the MOFM: When divided by the number of measurements, the discrete area enclosed by the target metric values and their previous minimal target metric values correspond to the OFM of the entire training process. The red arrow shows the cOFM and the black arrow is the mOFM. In this case, the target metric measures the cross-entropy loss of each fully trained target model on a Cifar10 validation set. (Right) The OFM plotted during training: We observe an increasing mismatch starting around epoch 600. Additionally, we show the range $(+, -)$ of the OFM and MOFM across a 5-fold cross-validation

## 5 Experimental setup

In our experiments, we focus on image-based self-supervised learning. However, it is likely that other target domains show mismatches as well, e.g., [13].

*Pretext tasks* For generation-based self-supervision, we evaluate the approaches of autoencoding data from autoencoders (CAE) and color restoration (CCAE) as suggested by Chen et al. [9]. To evaluate context structure generation we use denoising autoencoders (DCAE). Spatial context structure is evaluated via autoencoding transformations by predicting rotations [16] (RCAE). For context-based similarity methods, we follow the state-of-the-art contrastive learning approach from Chen et al. [9] (SCLCAE). We refer to the literature for first glances into mismatches for VAEs, meta-learning [41] and self-organization [58].

*Pretext models* Unless stated otherwise, we use a four-layer CNN as encoder. For the autoencoding data approaches, we use a four-layer decoder with transpose convolutions, for rotation prediction a single dense layer and for contrastive learning a nonlinear head, as suggested in [9]. We show that mismatches account for other architectures as well, by carrying out additional evaluations using ResNets [22] in Table 3 and Appendix C.

*Target tasks* We evaluate our metrics on image-based target tasks. For coarse-grained classification, we use Cifar10, Cifar100 [33] and the coarse-grained labels of 3dshapes [5]. For fine-grained classification, we use the PCam dataset [60] and the fine-grained labels of 3dshapes.

*Target models* Following the linear evaluation protocol, we use a single, linear dense layer (FC) as a target model with a softmax activation. To evaluate our metrics for other target models, we use a two MLP (2FC) and a three-layer MLP (3FC).

*Augmentations* We make sure not to compare augmentations instead of pretext tasks by following Chen et al. [9] for our base augmentations to which we add the pretext task-specific augmentations for pretext task training and evaluation. For the target task, we use the base training and evaluation augmentations of Chen et al. [9].

*Optimization* Our models are trained using the Adam optimizer [30] with standard parameters and batch size 2048 without any regularization instead of batch normalization. For our ResNets, we additionally use a weight decay of 1e−4.

*Mismatch evaluation* All reported values are determined by 5-fold cross-validation. We use standard early stopping (from tf.keras) as convergence criterion on the pretext evaluation curve with a minimum delta (threshold) of 0 and patience of 3. We change the patience in some experiments of Tables 1 and 3 to get a reasonable convergence epoch. For more details, we refer to Appendix B. When calculating our metrics, we estimate target values of missing epochs with linear interpolation to save computation time. In our case, SM3 and MM3 are measured on the target task accuracy.

*Implementation* Our implementation is available at https://github.com/BonifazStuhr/OFM.

## 6 Evaluation

In the following, we show results of most pretext and target tasks we have evaluated. We refer to Appendix C for additional, more detailed evidence. Since we capture our metrics during training, all mismatches are measured on the evaluation dataset.

**Table 1** MOFM, cSM3 and MM3 of the models from Fig. 4

| | CAE (Cifar10) | | DCAE (Cifar10) | | CCAE (Cifar100) | | CCAE (PCam) | | RCAE (PCam) | | | SCLCAE (3dshapes) | | |
|---|---|---|---|---|---|---|---|---|---|---|---|---|---|---|
| | cSM3 | MOFM | cSM3 | MOFM | cSM3 | MOFM | cSM3 | MOFM | cSM3 | MOFM | MM3 | cSM3 | MOFM | MM3 |
| **Rep. size** | | | | | | | | | | | | | | |
| 2x2x4 | **0.00** | **0.00** | 0.05 | **0.00** | 0.28 | 1.54 | 4.98 | 9.28 | 5.38 | 4.15 | −22.26 | 26.34 | ∞ | −7.99 |
| 2x2x32 | 0.07 | 1.99 | **0.00** | 3.20 | 0.65 | 3.64 | 5.17 | 34.30 | 3.34 | 7.92 | −21.09 | 12.98 | 36.39 | −57.67 |
| 2x2x128 | 0.20 | 10.10 | 0.06 | 5.51 | 0.51 | 0.81 | 0.32 | 0.10 | 1.03 | 4.04 | −23.47 | 8.14 | **22.65** | − **66.19** |
| 2x2x256 | 0.75 | 11.14 | 0.69 | 5.17 | 0.17 | **0.00** | 0.43 | 0.87 | 0.44 | **0.00** | −27.60 | 6.52 | 27.65 | −65.77 |
| 2x2x512 | 0.43 | 5.28 | 0.36 | 1.25 | **0.00** | **0.00** | 0.20 | 0.07 | 0.18 | **0.00** | − **28.03** | 5.96 | 27.78 | −63.61 |
| 2x2x1024 | 0.24 | **0.25** | **0.03** | **0.00** | **0.00** | **0.00** | **0.00** | **0.00** | **0.09** | **0.00** | −26.56 | **4.76** | 32.70 | −57.92 |
| **Target model** | | | | | | | | | | | | | | |
| FC | 0.75 | 11.14 | 0.69 | 5.17 | **0.00** | **0.00** | **0.00** | **0.00** | 1.03 | 4.04 | −23.47 | **6.52** | **27.65** | −65.77 |
| 2FC | **0.03** | 5.68 | **0.00** | 5.14 | 0.08 | **0.00** | **0.00** | **0.00** | **0.31** | 0.76 | −28.56 | 1.84 | 103.30 | −70.49 |
| 3FC | **0.03** | **3.94** | **0.00** | **3.17** | 0.12 | 0.02 | 0.08 | **0.00** | 0.37 | **0.61** | − **29.61** | **0.91** | 258.18 | − **71.26** |
| **Augmentations** | | | | | | | | | | | | | | |
| All | **0.75** | 11.14 | 0.69 | 5.17 | **0.17** | **0.00** | 0.43 | 0.87 | 1.03 | 4.04 | −23.47 | 6.52 | 27.65 | − **65.77** |
| NoIter | 0.99 | **10.99** | **0.50** | 2.33 | – | – | – | – | **0.58** | **0.10** | − **28.60** | **0.00** | 0.01 | −37.92 |
| NoIterNoFlip | 1.00 | 12.51 | 0.55 | **1.73** | – | – | – | – | 1.51 | 7.33 | −10.60 | **0.00** | **0.00** | −35.95 |
| NoFlip | – | – | – | – | 0.20 | **0.00** | **0.41** | **0.04** | – | – | – | – | – | – |

The smallest mismatches for each setup are printed in bold

MM3 is measured on the target and pretext task classification error. cSM3 is measured on the target task classification error and corresponds to the accuracy we would lose when we naively train the target model after pretext model convergence. Values are obtained by 5-fold cross-validation. We show the stability of each measurement in Tables 5, 6 and 7 of Appendix B

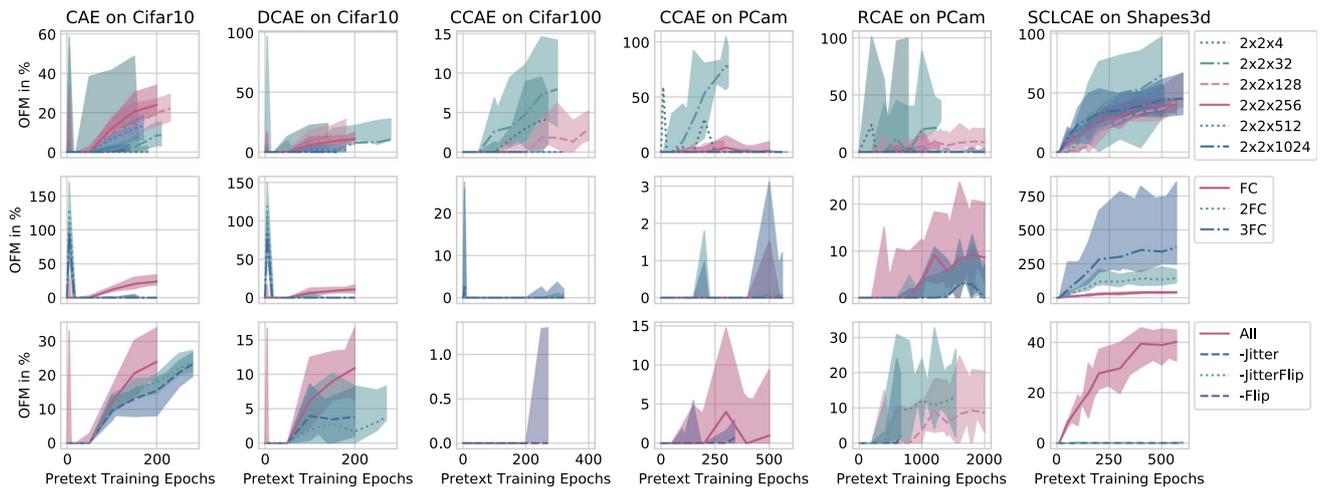

**Fig. 4** (Top) Impact of different pretext model representation sizes on the OFM. (Middle) OFM for the linear target model and nonlinear target models trained on our pretext model. (Bottom) OFM for the linear target model and for the pretext models trained on fewer augmentations. First, we removed the color jitter and then the vertical flip from the augmentations. The target models of SCLCAE were trained on 3dshapes to predict the object hue

## 6.1 Mismatch and convergence

For our measurements, we make sure to use metric value pairs from models that do not overfit. We achieve this by applying a convergence criterion on the pretext task and by using the best metric values from each target model evaluation curve. As shown in Appendix C, most observations in our experiments are independent of the use of a convergence criterion, if pretext models are trained long enough and without overfitting. Furthermore, we observe a common behavior in Fig. 1: Target models trained on higher epochs of the pretext model tend to converge faster. *This indicates that longer training of the pretext task tends to create easier separable representations which may mismatch with the class label.*

## 6.2 Stability

To evaluate the stability of our measurements, we show the mismatches of the entire training process and their range $(+, -)$ using 5-fold cross-validation in Figs. 2 and 3. The range of all other models, we have trained is shown in Appendix C. We observe that M3 generally seems more stable than SM3 or the OFM, since it does not rely so heavily on the target metric values, which can be quite unstable. The instability of the target task mismatch is captured in M3, but does not matter that much in the overall measurement for most cases. This is favourable if a stable value is desired and unfavourable if one wants to capture the instability of the target task training process explicitly. Furthermore, M3 is able to compensate target fluctuations with pretext fluctuations. In general, we observe that as long as we calculate the OFM across a fair amount of cross-validations (in our case 5), we can make statements about the mismatch. We measure our metrics on the mean losses during 5-fold cross-validation instead of calculating them five times and taking the average. For M3 both variants are equivalent and for the OFM measuring on the mean losses leads to a lower bound in the case where all models converge at step $s_n$ (see Appendix A for the simple proofs). We prefer to measure our metrics on the mean losses, since this avoids mismatches occurring just in some validation cycles due to small fluctuations of the underlying training procedure. An example is shown in translucent red in Fig. 3 at the beginning of training. We want to point out that the training and validation data differ slightly in every round because of the cross-validation setup. This increases the instability, but shows the general behavior of the metrics for the underlying data distribution. In Appendix C we compare the instability of partially measured mismatches using linear interpolation with mismatches measured for every pretext training epoch and observe a similar instability. However, when using the OFM in practice to compare models on a finer scale, we recommend to search for the actual minimal target metric value, since the OFM relies on this value at each step. However, when tuning a model for maximum performance, one searches for this value. Thereby looking at the OFM curve gives good indications in which interval one should search. This makes this protocol useful for performance tuning, if enough computational power is available.

## 6.3 Dependence on representation size

We hypothesize that large representation sizes tend to lower the OFM, which could be one reason why representation sizes are large in unsupervised learning. To affirm this hypothesis empirically, we train our pretext models with varying representation sizes on different target tasks while fixing all other model parameters. Figure 4 and Table 1 show that *the OFM tends to decrease when we enlarge the representation size*. A reason for that might be that target models can exploit the high dimensional space of large representations to find better fitting clusters for their target task. We found an exception of this behavior, where we use larger representations of SCLCAE for the easy task of object hue prediction. Here, the target models trained on the untrained pretext models with larger representation sizes already achieve high performance due to a larger number of color-selective, random features. Further learning of the pretext model, in this case, does not lead to a high-performance gain and forgetting these sensitive random features during training leads to a high mismatch. Additionally, we observe that mismatches decrease, when we decrease the representation size for generation-based methods. A reason could be that the pretext models are forced to generalize to solve the target task for small representation sizes due to the limited amount of features in the bottleneck, or simply underfit on the pretext task.

## 6.4 Dependence on target model complexity

In Fig. 4 and Table 1 we observe a OFM spike early in training for more complex target models. This spike occurs probably because nonlinear target models make better sense of specific random features at pretext task initialization, in contrast to the linear target model. *Besides early spikes, mismatches tend to decrease when we add complexity to the target model*. A model with increased nonlinearity has more freedom to disentangle representations, which do not fit properly with the target task. Again we found an exception where the MOFM is lower for linear models when predicting the object hue after contrastive learning which can be appointed to the color-selective, random features of the untrained pretext model.

## 6.5 Dependence on augmentations

We vary the augmentations used for the pretext and target model by removing the color jitter and the image flip from our base augmentations successively. Figure 4 shows that *augmentations can have a positive or negative impact on the mismatch*. E.g., when predicting the object hue, the ill-posed color jitter augmentation increases the mismatch significantly.

## 6.6 Dependence on target task type

Here we use our metrics to examine findings stated in [13, 32] and [64, 70], where it is argued that some pretext tasks are better suited for different target tasks. We fix the underlying data distribution by using the 3dshapes dataset and train our target models for the different tasks. These tasks require a generic understanding of the scene like coarse-grained knowledge about object type and hue and fine-grained knowledge about shapes, positions and scales. In Fig. 5 and Table 2, we observe that *pretext models tend to learn pretext task specific features and discard features that are not needed to solve the pretext task during training. These models, therefore, mismatch with ill-posed target tasks*. For example, rotation prediction discards features corresponding to the hue while it learns much about the orientation of the object.

## 6.7 Applying our metrics to ResNet models

In Fig. 6, we apply our metrics to ResNet models for several pretext and target tasks. For contrastive learning we

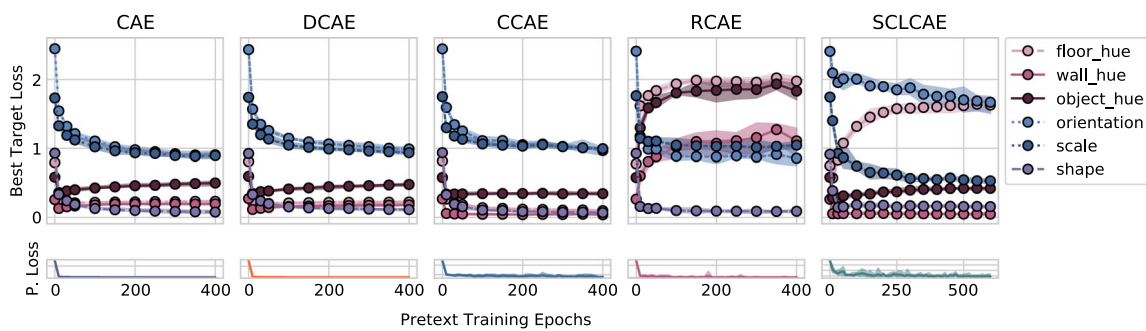

**Fig. 5** (Bottom) Pretext losses of our model trained for color restoration (CCAE), prediction rotations (RCAE), and contrastive learning (SCLCAE). (Top) Best target losses of linear models trained for the different prediction tasks of 3dshapes

**Table 2** cSM3, MOFM and MM3 on the 3dshapes dataset

| | CAE | | DCAE | | CCAE | | RCAE | | | SCLCAE | | |
|---|---|---|---|---|---|---|---|---|---|---|---|---|
| | cSM3 | MOFM | cSM3 | MOFM | cSM3 | MOFM | cSM3 | MOFM | MM3 | cSM3 | MOFM | MM3 |
| Floor_hue | 0.01 | 0.95 | **0.00** | 1.28 | **0.02** | **0.00** | 56.68 | $\infty$ | 44.67 | 28.18 | 268.27 | −48.38 |
| Wall_hue | 0.02 | 32.03 | 0.00 | 24.43 | 0.10 | **0.00** | 25.17 | $\infty$ | 7.80 | **0.29** | 0.46 | −**76.40** |
| Object_hue | 0.38 | 22.71 | 0.43 | 24.55 | 1.55 | 0.63 | 59.65 | $\infty$ | 40.1 | 2.87 | 8.69 | −73.17 |
| Scale | 0.41 | **0.00** | 0.27 | **0.00** | 0.10 | **0.00** | 2.60 | 0.13 | 31.78 | 2.43 | **0.00** | −44.80 |
| Shape | 0.07 | **0.00** | 0.08 | **0.00** | 0.03 | **0.00** | **0.20** | 0.06 | −**2.48** | 1.67 | 2.16 | −67.54 |
| Orientation | **0.00** | **0.00** | **0.00** | **0.00** | 0.23 | **0.00** | 0.48 | **0.00** | 22.26 | 2.50 | 6.68 | −9.11 |
| Average | 0.15 | 9.28 | **0.13** | 8.21 | 0.34 | **0.11** | 24.13 | $\infty$ | 24.02 | 6.32 | 47.71 | −**53.23** |

The smallest mismatches for each setup are printed in bold

SM3 and MM3 are measured on the target task accuracy. Values are obtained by 5-fold cross-validation. We show the stability of each measurement in Tables 8, 9 and 10 of Appendix B observe a small OFM for Cifar10 and Cifar100, which occurs late in training after pretext model convergence. However, when we use contrastive learning as pretext task for fine-grained tumor detection on the PCam dataset, we observe a mismatch before pretext model convergence. For the well-known rotation prediction pretext task, we oberved a high mismatch on Cifar10 classification early in pretext training. In Table 3, we show the corresponding mismatches measured until pretext model convergence.

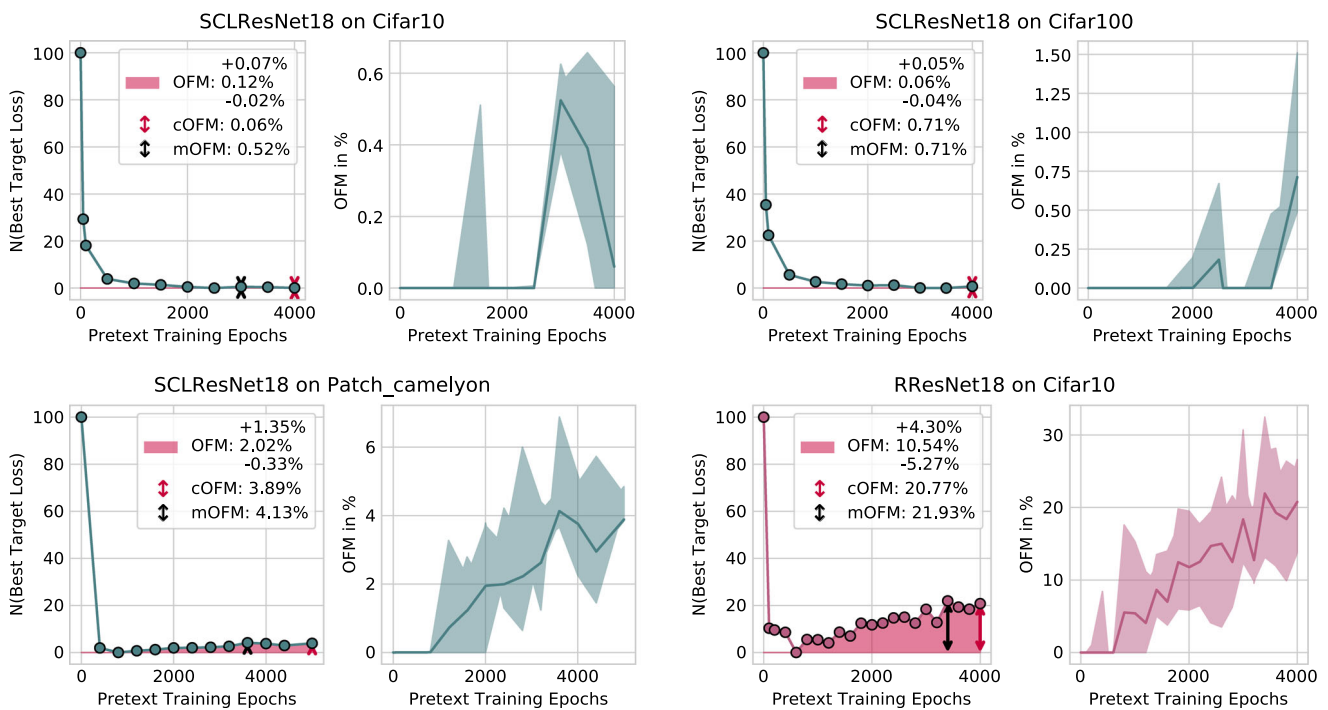

**Fig. 6** OFM for different pretext tasks trained with a ResNet18 model as a backbone. The mismatches are shown for the entire training process

# 7 Future work

In future work, our metrics can be used to create, tune and evaluate (self-supervised) representation learning methods for different target tasks and datasets. These metrics make it possible to quantify the extent to which a pretext task matches a target task, and to determine whether the pretext task learns the right kind of representation throughout the entire training process. This enables a comparison of methods on benchmarks across different pretext tasks and ACC stands for the best accuracy on the target task of all target models trained on the pretext model. cSM3 corresponds to the accuracy we would lose when we naively train the target model after pretext model convergence. Values are obtained by 5-fold cross-validation. The mismatches are measured with a convergence criterium models. The dependencies of the objective function mismatch on different parts of the self-supervised setup (e.g., representation size) can be explored by future work in more detail, to further evaluate our findings and to create pretext tasks and model architectures that are robust against mismatches. Our metrics are defined for setups where the target models are trained on pretext model representations in general. Therefore, they can also be applied to other representation learning areas such as supervised, semi-supervised, few-shot, or biological plausible representation learning.

**Table 3** Mismatches of ResNets with convergence criterium

|  | ResNet | | | |
|---|---|---|---|---|
|  | ACC | cSM3 | MOFM | MM3 |
| RCAE (Cifar10) | $54.64^{+1.80}_{-2.01}$ | $3.98^{+1.74}_{-3.60}$ | $4.87^{+4.42}_{-3.11}$ | $31.82^{+0.75}_{-0.68}$ |
| SCLCAE (PCam) | $96.25^{+0.44}_{-0.23}$ | $0.37^{+0.44}_{-0.37}$ | $0.86^{+1.00}_{-0.60}$ | $-53.26^{+0.52}_{-0.38}$ |

# 8 Conclusion

In this work, we have used the linear evaluation protocol as a basis to define and discuss metrics to measure the metrics mismatch and the objective function mismatch. With soft and hard versions of our metrics, we collected evidence of how these mismatches relate to the pretext model's representation size, target model complexity, pretext and target augmentations as well as pretext, and target task types. Furthermore, we observe that the epoch of the target task peak performance varies strongly for different datasets and pretext tasks. This highlights the importance of the protocol and shows that comparing approaches after a fixed number of epochs does not yield the entire picture of their capability. Our protocols make it possible to define benchmarks across different target tasks, where the goal is not to mismatch with the target metrics while achieving the best possible performance.

**Supplementary Information** The online version contains supplementary material available at https://doi.org/10.1007/s00521-022-07031-9.

**Acknowledgements** Really kindly, we want to thank our colleagues from the University of Applied Sciences Kempten and the Autonomous University of Barcelona for the helpful discussions about this topic. Especially we want to thank Markus Klenk for proofreading the draft and Jordi Gonzàlez for his much-appreciated feedback on this work.

**Funding** Open Access Funding provided by Universitat Autonoma de Barcelona.

**Data availability** All datasets are available in the public domain.

**Code availability** Our implementation is available at https://github.com/BonifazStuhr/OFM.

## Declarations